\documentclass[10pt,twocolumn,letterpaper]{article}

\usepackage{iccv}
\usepackage[utf8x]{inputenc}
\usepackage{times}
\usepackage{epsfig}
\usepackage{graphicx}
\usepackage{amsmath}
\usepackage{amssymb}

\usepackage{tabularx}
\usepackage{multirow}

\usepackage{subfigure}
\usepackage{algorithm}
\usepackage{algorithmic}
\usepackage{float}
\usepackage{bm}

\usepackage[breaklinks=true,bookmarks=false]{hyperref}

\iccvfinalcopy %

\def\iccvPaperID{2070} %

\ificcvfinal\fi
\begin{document}

\title{Universal Perturbation Attack Against Image Retrieval}

\author{
    Jie Li$^{1}$,
    Rongrong Ji$^{1,2}$\thanks{Corresponding author.},
    Hong Liu$^{1\*}$\*,
    Xiaopeng Hong$^{3,2,4}$,
    Yue Gao$^{5}$,
    Qi Tian$^{6}$
    \\
    $^{1}$Department of Artificial Intelligence, School of Informatics, Xiamen University, \\
    $^{2}$Peng Cheng Lab, Shenzhen, China,  \\
    $^{3}$MOE Key Lab. for Intelligent Networks and Network Security/Faculty \\
    of Electronic and Information Engineering, Xi'an Jiaotong University, PRC \\
    $^{4}$University of Oulu, Finland,
    $^{5}$Tsinghua University,
    $^{6}$Huawei Noah’s Ark Lab
    \\
    {\tt\small lijie32@stu.xmu.edu.cn,}
    {\tt\small rrji@xmu.edu.cn,}
    {\tt\small lynnliu.xmu@gmail.com,} \\
    {\tt\small hongxiaopeng@mail.xjtu.edu.cn,}
    {\tt\small kevin.gaoy@gmail.com,}
    {\tt\small tian.qi1@huawei.com,}
}

\maketitle

\begin{abstract}

Universal adversarial perturbations (UAPs), a.k.a. input-agnostic perturbations, has been proved to exist and be able to fool cutting-edge deep learning models on most of the data samples.
Existing UAP methods mainly focus on attacking image classification models.
Nevertheless, little attention has been paid to attacking image retrieval systems.
In this paper, we make the first attempt in attacking image retrieval systems.
Concretely, image retrieval attack is to make the retrieval system return irrelevant images to the query at the top ranking list. 
It plays an important role to corrupt the neighbourhood relationships among features in image retrieval attack.
To this end, we propose a novel method to generate retrieval-against UAP to break the neighbourhood relationships of image features
via degrading the corresponding ranking metric.
To expand the attack method to scenarios with varying input sizes or untouchable network parameters,
a multi-scale random resizing scheme and a ranking distillation strategy are proposed.
We evaluate the proposed method on four widely-used image retrieval datasets, and report a significant performance drop in terms of different metrics, such as mAP and mP@10.
Finally, we test our attack methods on the real-world visual search engine, i.e., Google Images, which demonstrates the practical potentials of our methods.
\end{abstract}

\section{Introduction}
\begin{figure}[!t]
\centering
\includegraphics[width=3.3in]{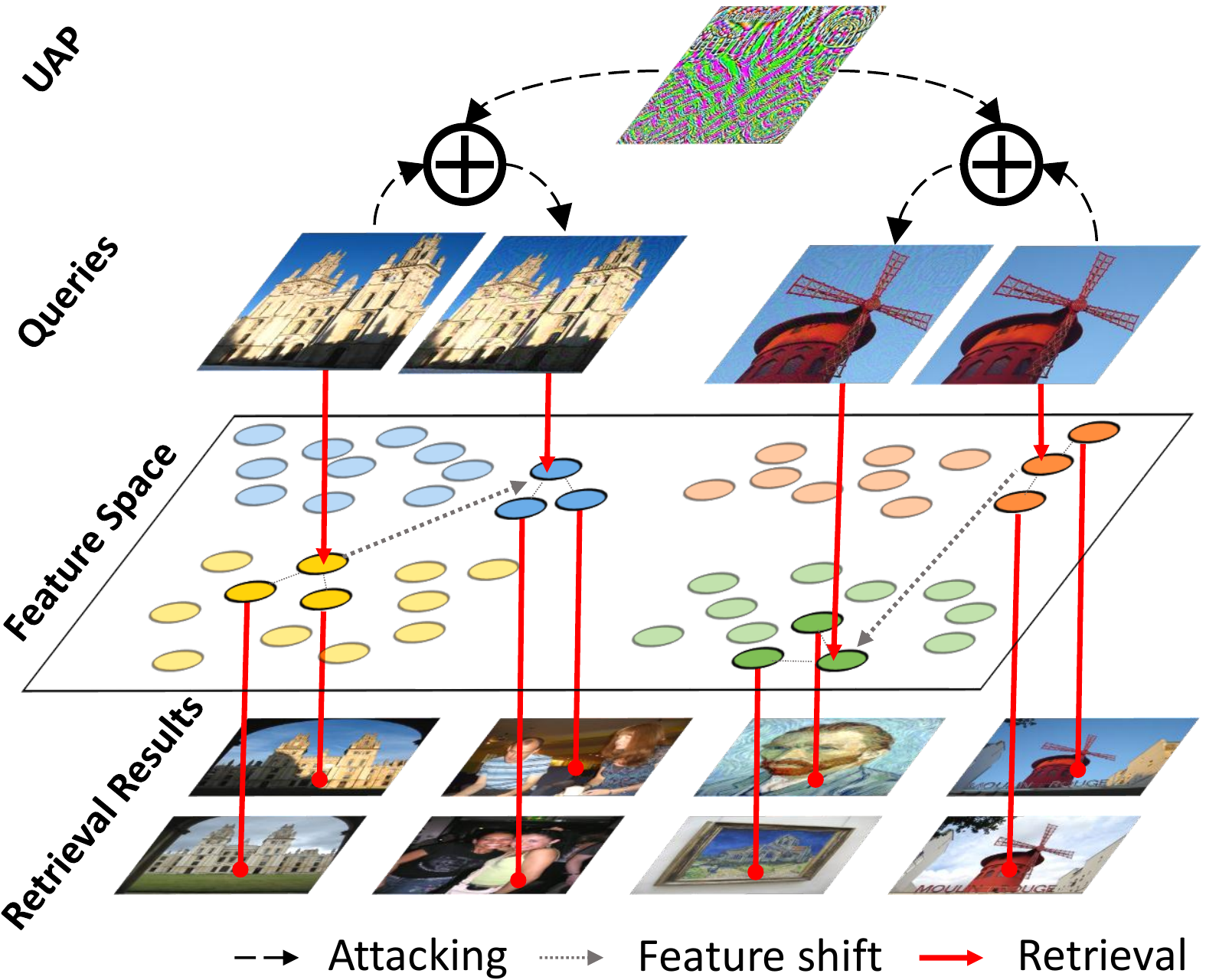}
\caption{When added to natural images, a single universal perturbation that is invisible to human eyes causes most images to shift significantly in the feature space without preserving original neighbourhood relationships. The top is the perturbation and dots represent the features of images. (Best viewed in color.)}
~\label{fg_introduction}
\vspace{-2em}
\end{figure}

Convolutional neural networks (CNN) have been the state-of-the-art solution for a wide range of computer vision tasks, such as image classification, image segmentation and objective detection. 
Despite the remarkable success, deep learning models have shown to be vulnerable to small perturbations to the input image. 
Various attack techniques have been proposed, like model distillation~\cite{Papernot:2017fk, xiao2018generating}, transfer learning~\cite{liu2016delving, Zhou_2018_ECCV}, and gradient updating~\cite{suvery1}.
In contrast to previous methods called image-specific perturbations having to perform computation every time to generate a particular perturbation for any given image,
Moosavi-Dezfooli \emph{et al.}~\cite{MoosaviDezfooli:2017ud} proposed an image-agnostic perturbation termed universal adversarial perturbation (UAP), which can fool most images from a data distribution. 
Being universal, UAPs can be conveniently exploited to perturb unseen datapoints on-the-fly without extra computation.
Therefore, UAPs are particularly useful in a wide range of applications.

However, existing methods regardless of whether image-agnostic or not, mainly focus on image classification while no existing work has touched the topic of attacking image retrieval systems.
As a long-standing research topic in computer vision~\cite{zheng2018sift}, image retrieval aims to find relevant images from a dataset given a query image.
Despite the extensive efforts in improving the search accuracy (\emph{e.g.}, new features like NetVLAD~\cite{arandjelovic2016netvlad} and generalized-mean pooling~\cite{radenovic2018fine}) or efficiency (\emph{e.g.}, indexing schemes like Hamming Embedding~\cite{jegou2008hamming} or hashing~\cite{liu2018ordinal, wang2018survey}), very little attention has been paid to the vulnerability of the state-of-the-art retrieval systems.
It is difficult, or even infeasible to apply existing UAPs methods in image retrieval directly.
The reasons come from four aspects.

\begin{itemize}
\item
    Different dataset label formats.
    Most existing UAP methods designed for image classification work on datasets labeled by categories~\cite{deng2009imagenet}, which need UAPs pushing datapoints across decision boundary~\cite{MoosaviDezfooli:2017ud}.
    However, datasets in retrieval are usually labeled by similarity~\cite{radenovic2018revisiting}, which require UAPs to capture complex relationships among features instead.

\item 
    Different goals.
    The goal of existing UAP methods is to disturb unary and binary model outputs for single instance, \emph{e.g.,} to change the most likely predict label.
    However, merely corrupting the top-$1$ result is still not enough since the retrieval evaluation is usually done on a ranking list.
Thus, to attack retrieval systems, one should disturb the ranking list via lowering the positions of positive samples there.

\item 
    Different sizes of model input.
    Generally, models which existing UAPs trained on ask for fixed-size input images, accordingly the size of UAPs is fixed as the input.
    However, these UAPs are fragile and can be defensed by varying the size of input~\cite{xie2017mitigating}.
    Note that, the size of images in retrieval usually vary, which restricts the direct usage of the traditional UAPs
    and thus poses a higher demand for generating UAP for the task of image retrieval.

\item
    Different model output and optimization methods.
    It is often assumed predict confidence of each category can be fetched~\cite{bhagoji2018black, chen2017zoo}, and the confidences are a group of continuous and floating numbers responding to the changes of input rapidly.
    It indicates a way to estimate gradient for optimization.
    However, the large-scale discrete ranking list returned by retrieval systems offers little guidance on approximating gradient.
    This fact makes it infeasible to apply existing UAPs to retrieval systems with network parameters unaccessible.

\end{itemize}

In this paper, we make the first attempt in attacking image retrieval, especially the cutting-edge image retrieval model that are deployed upon deep features.
In principle, 
we aim to generate a UAP for corrupting neighbourhood relationships in the feature space 
as depicted in Fig.\,\ref{fg_introduction}.
To address the challenges mentioned above, we propose a novel universal adversarial perturbation attack method for image retrieval.
\begin{figure}[!t]
\centering
\includegraphics[width=1.0\linewidth]{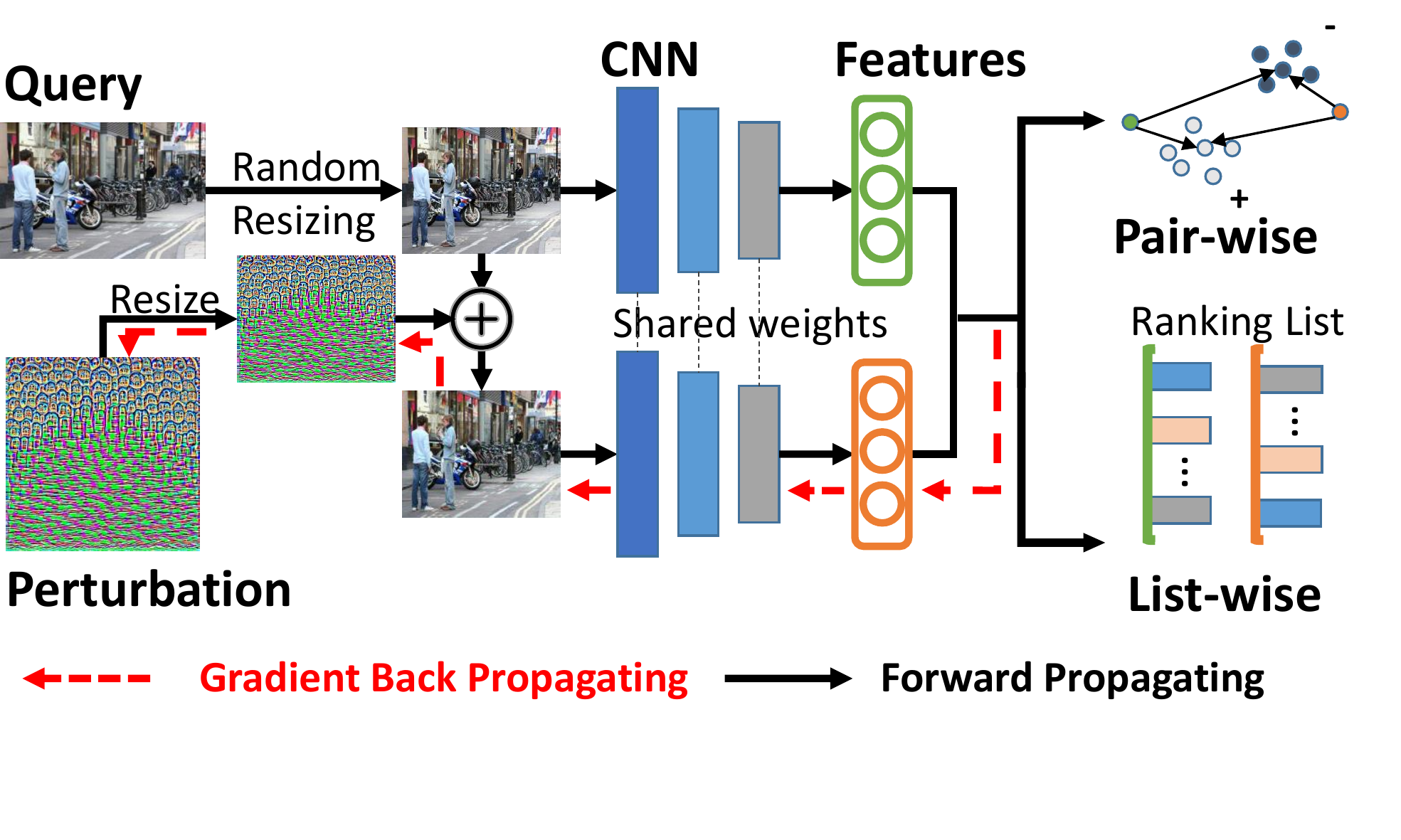}
\vspace{-3em}
\caption{The pipeline of the proposed method. Perturbation is first resized to the same size of the input image which goes through random resizing layer with a random scale. Then both the resized input image and the sum of perturbation and input image are fed into CNN model to corrupt three relationships. Only gradient of the perturbation will be calculated during back propagation to update the perturbation.
}\label{fg_framework}
\vspace{-1em}
\end{figure}
In detail, we build a general model to craft the UAP that breaks the neighborhood relationships among feature points by altering the input slightly.
Pair-wise relationship among neighborhood structures is first considered via constructing tuples based on the nearest and farthest groups. 
We corrupt this relationship by swapping the similarity relationship in the tuples.
Although corrupting pair-wise relationship is simple and efficient, the pair-wise information focuses on the local relationship between query and two data samples each time without considering global ranking list which is more significant for retrieval.
We argue it can not solve the retrieval attack problem fundamentally.
Eventually, we propose the approach to generate UAPs from list-wise aspect that goes further to permutate the entire ranking list via destroying the corresponding ranking metrics to lower positions of relative references.
In addition, we propose a multi-scale random resizing scheme to apply UAP to input images at different resolutions, which shows better attack performance than fixed-scale methods experimentally.
The pipeline of the proposed method is shown in Fig.\,\ref{fg_framework}. 

\begin{figure}[!t]
\centering
\includegraphics[width=1.0\linewidth]{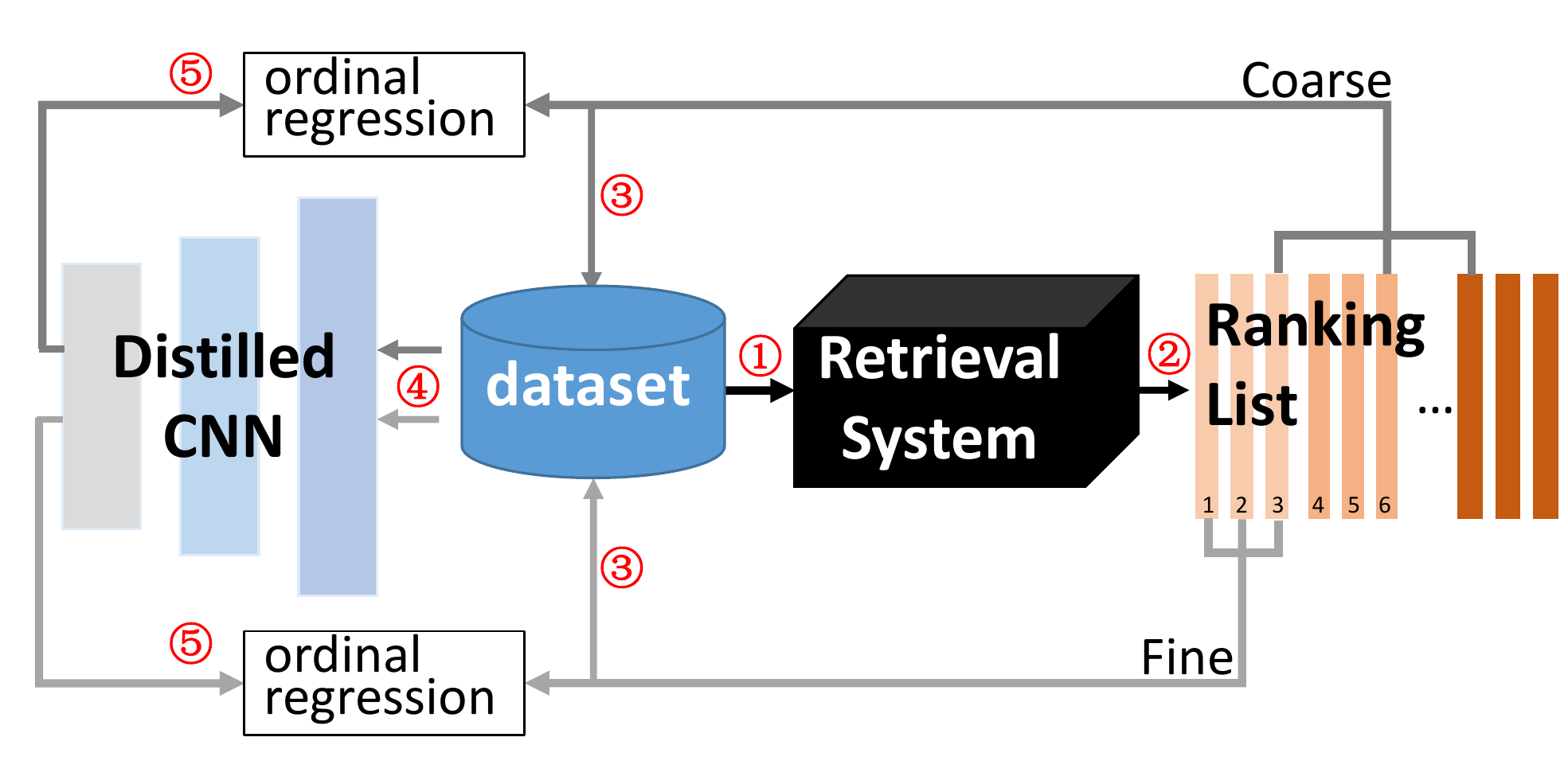}
\vspace{-2em}
\caption{
    The pipeline of ranking distilling.
    First, ranking list of the unknowable black-box retrieval systems for the dataset is obtained and divided into groups (\raisebox{.5pt}{\textcircled{\raisebox{-.9pt} {1}}}\raisebox{.5pt}{\textcircled{\raisebox{-.9pt} {2}}}).
    Then the datapoints of the coarse-grained subset randomly sampled from each group and the fine-grained top-$k$ references are fetched from dataset (\raisebox{.5pt}{\textcircled{\raisebox{-.9pt} {3}}}) to optimize the distilled model via regressing ordinal information among them (\raisebox{.5pt}{\textcircled{\raisebox{-.9pt} {4}}}\raisebox{.5pt}{\textcircled{\raisebox{-.9pt} {5}}}).
}\label{fg_distillation}
\vspace{-1em}
\end{figure}

Our scheme further enables attack without touching network parameters via a coarse-to-fine strategy to distill victim model by regressing ranking list as depicted in Fig.\,\ref{fg_distillation}. 
First, we construct coarse-grained subsets which preserves global ranking information sampled from the entire large-scale ranking list, and prompt distilled model to fit the ordinal relation in the subsets.
Then from the fine-grained level, we focus on the top-$k$ most related instances for retrieval to refine the distilled model.

The proposed method achieves high attack performance substantially and leads to a large performance drop on standard image retrieval benchmarks, \emph{i.e.}, Oxford Buildings and Paris with their revised versions.
The retrieval performance is tested on two CNN-based image representation~\cite{radenovic2018fine, razavian2016visual, tolias2015particular} with three different CNN models~\cite{he2016deep, krizhevsky2012imagenet, simonyan2014very}.
Quantitatively, the universal adversarial perturbation can drop the performance such as \emph{m}AP and \emph{m}P@10 by at least $50\%$, which reveals the cutting-edge image retrieval systems is quite vulnerable to adversarial examples. 
Interestingly, we further evaluate our universal perturbation on the real-world image search engine, \emph{i.e.}, Google Images, and conclude that the perturbation can also corrupt the output ranking list.

\section{Related Work}\label{sec:R}

\textbf{Adversarial Examples.}
Szegedy \emph{et al.}~\cite{szegedy2013intriguing} have demonstrated that neural networks can be fooled by adversarial example, which is a clean image being intentionally perturbed, \emph{e.g.} by adding adversarial perturbation that is quasi-imperceptible to human eyes.
Subsequently, various methods have been proposed to generate such perturbations~\cite{dong2018boosting, goodfellow6572explaining, moosavi2016deepfool}.
An iterative scheme is proposed in~\cite{kurakin2016adversarial} to achieve better attack performance via applying gradient ascent multiple times.
Besides, complex approaches like~\cite{moosavi2016deepfool} find perturbation from the perspective of classification boundary.

However, these methods compute perturbations for each data point specifically and independently. 
More recently, Moosavi-Dezfooli \emph{et al.}~\cite{MoosaviDezfooli:2017ud} have shown that there exists a single image-agnostic perturbation termed universal adversarial perturbation (UAP) being able to corrupt most natural images.
UAP is a single adversarial noise that is trained offline and can perturb the corresponding outputs of a given model online.
Contrast to white-box attack where victim model can be accessed, black-box attack refers to the case that attackers have little knowledge about victim.
It is observed that perturbations crafted for specific models or training sets can fool other models and datasets~\cite{goodfellow6572explaining, szegedy2013intriguing}, referred as transfer attack, which is widely adopted in black-box.
Another popular method is knowledge distillation~\cite{hinton2015distilling}, which obtains substitute model via regressing output of victim and then applies white-box attack methods~\cite{xiao2018generating}.

\textbf{Visual Features for Retrieval.}
Image retrieval is a long-standing research topic in computer vision~\cite{zheng2018sift}.
Given a query image, the search engine retrieves related ones from a large set of reference images.
A typical setting refers to extracting and comparing features between a query and references, such as global descriptors~\cite{oliva2001modeling} and local descriptor aggregations~\cite{jegou2010aggregating, sanchez2013image}.
Nowadays, the most prominent retrieval methods are mostly based on CNNs~\cite{babenko2015aggregating, babenko2014neural, hyeonwoo2017large, kalantidis2016cross,  radenovic2018fine, razavian2016visual, tolias2015particular}.
They mainly use the pre-trained CNNs as a backbone to extract global representation for images.
To that effect, CNN models pre-trained with ImageNet~\cite{deng2009imagenet} (\eg{} AlexNet~\cite{krizhevsky2012imagenet}, VGGNet~\cite{simonyan2014very} and ResNet~\cite{he2016deep}) already provide superior performance over hand-crafted features~\cite{babenko2014neural}.
Babenko \emph{et.al.}~\cite{babenko2014neural} further showed that fine-tuning the CNN models can further boost the retrieval performance.
In this trend, many recent methods are proposed to construct trainable pooling layers for better feature representation. 
Representative methods include, but not limited to, maximum activations of convolutions (MAC)~\cite{razavian2016visual, tolias2015particular}, weighted sum pooling (CroW)~\cite{kalantidis2016cross}, and generalized-mean pooling (GeM)~\cite{radenovic2018fine}.
In this paper, we mainly consider two state-of-the-art pooling methods, \emph{i.e.}, MAC~\cite{razavian2016visual, tolias2015particular} and GeM~\cite{radenovic2018fine}, with three different CNN models, \emph{i.e.}, AlexNet, VGGNet, and ResNet, to evaluate the performance of UAP attack.

\section{The Proposed Method}\label{sec:Method}
Our method is aimed to seek a universal perturbation $\delta$ with constraint of $\|\delta\|_\infty\leq  \epsilon$, to corrupt as much similarity relationships as possible in the data distribution $\mathcal{X}$.
By doing so, the originally similar features should be dissimilar after adding a small perturbation.

For convenience, we denote the universal perturbation by $\delta$, denote the feature vectors of the $i$-th original image $x_i$ and the adversarial one by:
\begin{align}
    f_i &=  F\big(R_I(x_i)\big), \nonumber \\
    f_i^{'} &= F\Bigg(\max \bigg(0, \min \Big(255,R_P \big(\delta, R_I(x_i)\big)+R_I(x_i) \Big) \bigg) \Bigg), \nonumber
\end{align}
where $F(\cdot)$ is the function that outputs the feature vector through a CNN model, $R_I(\cdot)$ and $R_P(\cdot, \cdot)$ are resizing operators for input image and universal perturbation, respectively. 
Resizing operators will be elaborated in Sec.\,\ref{resizing}.
The Euclidean distance between two feature vectors $f_i$ and $f_j$ are characterized as a function $d(f_i, f_j)$.
To avoid computational overhead caused by the large-scale dataset, a landmark-based ordinal relation~\cite{arias2015some} that compares any query point to the landmarks\footnote{Landmarks are generated via K-means clustering.} is calculated in advance.

\subsection{Baseline}
We first attempt to disturb retrieval systems using label-wise information to validate whether UAPs against classification is suitable for retrieval.
We define a classifier, equipped by the cross-entropy loss function with FC layer and softmax layer.
We recognize the cluster index as pseudo-label and use them for all experiments in order to reduce computational cost and to ensure every experiment is conducted under the same setting.
Pseudo-labels are crafted with features from victim model, which include more victim attributes than exact labels and could benefit attack.
Furthermore, pseudo-labels can be easily extended to scenarios where exact labels are unavailable.
The classifier is trained via
pseudo-labels
and then fooled by minimizing the widely-used classification attack loss~\cite{carlini2017towards} as follows:
\begin{equation}
    L(\delta) = {[{Z(x')}_t - \max\big({Z(x')}_i:i \neq t\big)]}_{+},
\end{equation}
where ${[x]}_{+}$ is the $\max(x, 0)$ function, $Z(\cdot)$ is the output before the softmax layer, $t$ is the label of clean input and $x'$ is the input perturbed by $\delta$.

\subsection{The Proposed UAP}\label{sub_sec:white_methd}

\begin{algorithm}[!t]
\caption{Universal Perturbation Generating for Attacking Image Retrieval.}\label{alg1}
\renewcommand{\algorithmicrequire}{\textbf{Input:}}
\renewcommand{\algorithmicensure}{\textbf{Output:}}
\begin{algorithmic}[1]
    \REQUIRE{}
Data set $X=\{x_1, x_2, \ldots, x_n\}$, parameters $\lambda$.
\ENSURE{}
Universal perturbation vector $\delta$.
\STATE{} Initialize $\delta \leftarrow 0$
\REPEAT{}
\FOR{each datapoint $x_i \in X$}
\STATE{Randomly resize $x_i$ then resize perturbation $\delta$ accordingly}
\STATE{Compute and update the gradients $\nabla L$}
\STATE{Update the perturbation by optimizing Eq.\,\ref{Eq_up}}
\IF{$\delta$ gets saturated}
\STATE{$\delta = \delta / 2$}
\ENDIF{}
\ENDFOR{}
\UNTIL{convergence.}
\end{algorithmic}
\end{algorithm}

Image retrieval can be viewed as a ranking problem, from which perspective the relationship between query and references plays an important role~\cite{liu2009learning}.
Therefore, such relationship should be fully utilized, that can further improve the attack performance.
To this end, we consider two relationships to be corrupted \emph{i.e.}, pair-wise and list-wise. 

\textbf{Corrupting Pair-wise Relationship.}\label{method:pair}
Here we use ordinal relationship between the nearest and the farthest references to approximate the pair-wise information, which can be constructed directly via the classical triplet loss.
Formally, an ordered relation set $C$ can be written as follows:
    \begin{align}
    & \small{\eta_{ij}<\eta_{ik} \Rightarrow d(f_j, f_i) > d(f_k, f_i)}  \nonumber
      \\
    & \small{\Rightarrow d(f_j, f_i^{'}) < d(f_k, f_i^{'}), \forall(i,j,k)\in C.}
    \end{align}
We define $\eta_{ik} = 1$ as similar pairs of $x_i$ and $x_k$ that share the same cluster.
$\eta_{ij} = 0$ means the distance between clusters corresponding to samples $x_i$ and $x_j$ is the farthest.
Therefore, a set of tuples belonging to the subset of $C$ can be re-computed.
To attack the retrieval system, we minimize the traditional triplet loss as follows:
\begin{equation}\label{eq_pair_loss}
    L(\delta)= \sum\nolimits_{\eta_{ik}=1, \eta_{ij}=0}{[\alpha + d_1(f_j, f_i^{'}) - d_1(f_k, f_i^{'})]}_+,
\end{equation}
where $\alpha$ is  the parameter representing the margin between the matched and unmatched samples.

\textbf{Corrupting List-wise Relationship.}\label{method:list}
Unlike corrupting pair-wise one that focuses merely on the local relationship, we further permutate the entire ranking list for list-wise relationship to destroy the corresponding ranking metric.

Since the list is typically too large to be directly processed, we re-use the landmark employed above and construct a subset of the ranking list with suitable size by sampling references from each landmark each time.
We treat the reversed ranking list of cluster centers as the ideal ranking sequence, and destroy the normalized Discounted Cumulative Gain (NDCG) metric~\cite{jarvelin2000ir} as it is the most classical measurement well suited to information retrieval~\cite{robertson2007rank}.
NDCG is multilevel measures, which is aimed to measure the instance's gain based on its position in result list. 
The gain is accumulated from the top of the list to the bottom, and gain of reference at lower rank will be discounted.
Given any permutation $g$ of the set $S$ and its ratings sets ${\{y_i\}}_{i=1}^{|g|}$, DCG is defined as follows:
\begin{equation}\label{eq_ndcg}
    DCG(R) = \sum_{i=1}^{|g|} \frac{2^{y_i} - 1}{\log_2 (i+1)}.
\end{equation}
NDCG divides DCG by value of ideal ranking sequence to ensure a range of $[0,1]$.

However, the function in Eq.\,(\ref{eq_ndcg}) is non-convex and non-smooth, which makes the optimization problematic.
To this end, we approximate the gradient by accumulating the influence via swapping references.
After sorting the images by score for a given query image feature $f_i$, if $y_j$ and $y_k$ are the ideal rank indices of current the $i$-th and $j$-th images feature $f_j$ and $f_k$ respectively, we have the tangent of the distance function that has the property as follows:
\begin{align}\label{eq_reG}
  \frac{\partial d(f_i,f_j)}{\partial \delta} - \frac{\partial d(f_i,f_k)}{\partial \delta} \gg 0, \nonumber
  \\
    ~\text{whenever}~j\gg k ~\text{and}~ y_j \ll y_k.
\end{align}
Therefore, given a ranking list, we can directly calculate the sum of the gradient residuals in Eq.\,(\ref{eq_reG}), which roughly approximate the gradient of the NDCG loss in Eq.\,(\ref{eq_ndcg}).
Due to the discounted factor in DCG, following the similar strategy in~\cite{burges2007learning}, we also introduce the $\lambda$ parameter to weight the gradient residual, whose gradient can be defined as follows:
\begin{align}
& \nabla \delta = \frac{\partial NDCG(R)}{\partial \delta} \approx \sum_{j \neq k} \lambda_{jk} (\frac{\partial d(f,f_j)}{\partial \delta}-\frac{\partial d(f,f_j)}{\partial \delta}), \nonumber\\
&\lambda_{jk} = \frac{-1}{1+e^{(d(f,f_j)-d(f,f_k))}} |{\Delta_{NDCG}}_{jk}|,
\end{align}
where $|{\Delta_{NDCG}}_{ij}|$ is the change of NDCG metric if swap positions of the $i$-th and the $j$-th references.

\begin{table*}[!t]
\scalebox{0.76}[0.8]{\begin{tabular}{c|l|cccc|ccc|cccc|ccc|c}
\hline
 &  & \multicolumn{1}{c|}{\multirow{2}{*}{Oxford5k}} & \multicolumn{6}{c|}{$\mathcal{R}$Oxford5k} & \multicolumn{1}{c|}{\multirow{2}{*}{Paris6k}} & \multicolumn{6}{c|}{$\mathcal{R}$Paris6k} &  \\ \cline{1-2} \cline{4-9} \cline{11-17} 
 &  & \multicolumn{1}{c|}{} & \multicolumn{1}{c|}{E} & \multicolumn{1}{c|}{M} & H & \multicolumn{1}{c|}{E} & \multicolumn{1}{c|}{M} & H & \multicolumn{1}{c|}{} & \multicolumn{1}{c|}{E} & \multicolumn{1}{c|}{M} & H & \multicolumn{1}{c|}{E} & \multicolumn{1}{c|}{M} & H &  \\ \hline

Eval &  &  & mAP &  &  &  & mP@10 &  &  & mAP &  &  &  & mP@10 &  & mDR \\ \hline

\multirow{4}{*}{A-MAC} 
 & O & 57.11 & 45.23 & 32.96 & 10.43 & 57.25 & 55.43 & 15.36 & 65.64 & 63.99 & 46.93 & 20.06 & 88.00 & 91.29 & 58.29 &  \\
 & C & 46.99 & 36.13 & 27.89 & 7.86 & 49.58 & 48.36 & 12.71 & 57.91 & 52.96 & 40.33 & 16.27 & 80.86 & 83.00 & 48.86 & 15.47\% \\
 & P & 29.61 & 24.52 & 17.99 & 4.92 & 32.06 & 30.86 & \textbf{6.67} & 42.89 & 38.71 & 30.43 & 11.13 & 52.86 & 54.71 & 29.14 & 44.35\% \\
 & L & \textbf{27.88} & \textbf{21.59} & \textbf{16.31} & \textbf{4.06} & \textbf{28.33} & \textbf{28.57} & 7.50 & \textbf{41.15} & \textbf{37.40} & \textbf{29.28} & \textbf{10.00} & \textbf{49.29} & \textbf{51.43} & \textbf{25.00} & \textbf{48.33}\%
 \\ \hline

\multirow{4}{*}{A-GeM} 
 & O & 59.86 & 50.21 & 36.72 & 14.29 & 58.10 & 53.60 & 23.32 & 73.66 & 70.65 & 51.89 & 22.80 & 87.71 & 88.86 & 57.86 &  \\
 & C & 35.49 & 30.07 & 22.00 & 7.03 & 33.62 & 31.71 & 10.16 & 48.27 & 42.60 & 33.80 & 12.55 & 46.57 & 50.00 & 27.00 & 43.51\% \\
 & P & 29.31 & 22.85 & 17.57 & 5.56 & \textbf{25.65} & 24.79 & 8.36 & 40.71 & 35.17 & 29.44 & 10.71 & 38.86 & 41.71 & 20.14 & 54.12\% \\
 & L & \textbf{26.48} & \textbf{22.45} & \textbf{17.12} & \textbf{5.29} & 25.78 & \textbf{24.25} & \textbf{8.03} & \textbf{37.17} & \textbf{32.28} & \textbf{27.42} & \textbf{10.23} & \textbf{34.86} & \textbf{37.14} & \textbf{18.29} & \textbf{56.88}\%
 \\ \hline

\multirow{5}{*}{V-MAC} 
 & O & 81.45 & 75.07 & 57.15 & 29.96 & 78.60 & 78.33 & 45.57 & 88.31 & 86.39 & 69.60 & 44.97 & 93.57 & 96.86 & 84.71 &  \\
 & C & 42.70 & 37.15 & 30.14 & 14.87 & 35.59 & 36.14 & 20.43 & 34.15 & 29.88 & 27.37 & 12.48 & 18.57 & 18.86 & 12.43 & 61.80\% \\
 & P & 37.60 & 32.33 & 26.99 & 14.49 & 35.15 & 35.29 & 20.57 & \textbf{23.76} & \textbf{21.02} & \textbf{20.12} & \textbf{9.21} & \textbf{13.86} & \textbf{15.57} & \textbf{9.86} & 66.94\% \\
 & L & \textbf{35.57} & \textbf{29.83} & \textbf{24.97} & \textbf{13.13} & \textbf{32.79} & \textbf{32.29} & \textbf{19.71} & 25.38 & 22.13 & 20.99 & 9.23 & 15.29 & 17.14 & 10.43 & \textbf{67.96}\%
 \\ \hline

\multirow{4}{*}{V-GeM} 
 & O & 85.24 & 76.43 & 59.17 & 32.26 & 80.52 & 81.29 & 49.71 & 86.28 & 84.66 & 67.06 & 42.40 & 95.14 & 97.57 & 83.00 &  \\
 & C & 46.08 & 38.98 & 31.59 & 14.20 & 36.45 & 36.29 & 19.57 & 44.51 & 38.05 & 34.44 & 15.39 & 27.14 & 27.29 & 17.57 & 57.60\% \\
 & P & 43.71 & 37.84 & 30.92 & \textbf{15.36} & 36.76 & 37.00 & 21.86 & \textbf{30.92} & 28.12 & \textbf{25.78} & 11.91 & 17.43 & 17.43 & 12.86 & 62.64\% \\
 & L & \textbf{41.94} & \textbf{37.13} & \textbf{30.00} & 15.39 & \textbf{34.40} & \textbf{34.00} & \textbf{21.43} & 32.29 & \textbf{27.39} & 25.95 & \textbf{11.69} & \textbf{16.86} & \textbf{16.86} & \textbf{10.86} & \textbf{63.72}\%
 \\ \hline

\multirow{4}{*}{R-MAC} 
 & O & 81.69 & 73.85 & 56.14 & 29.80 & 78.33 & 79.86 & 46.57 & 83.55 & 81.56 & 63.91 & 39.06 & 93.52 & 96.71 & 79.57 &  \\
 & C & 58.52 & 50.65 & 37.50 & 15.59 & 56.47 & 54.29 & 24.71 & 67.57 & 61.51 & 49.43 & 25.01 & 70.00 & 72.43 & 49.57 & 31.27\% \\
 & P & 35.31 & 30.34 & 24.73 & 13.37 & 36.62 & 36.43 & 20.71 & 35.66 & 32.61 & 27.23 & 12.12 & 32.57 & 34.86 & 21.29 & 59.71\% \\
 & L & \textbf{34.08} & \textbf{28.68} & \textbf{23.30} & \textbf{12.09} & \textbf{34.26} & \textbf{32.95} & \textbf{19.86} & \textbf{34.63} & \textbf{30.71} & \textbf{26.16} & \textbf{11.50} & \textbf{28.00} & \textbf{29.71} & \textbf{18.43} & \textbf{62.60}\%
 \\ \hline

\multirow{4}{*}{R-GeM} 
 & O & 86.24 & 80.63 & 63.13 & 38.51 & 82.72 & 83.14 & 54.57 & 90.66 & 90.33 & 74.06 & 51.69 & 94.96 & 98.29 & 88.29 &  \\
 & C & 68.45 & 59.30 & 45.57 & 21.38 & 66.25 & 62.52 & 34.86 & 79.00 & 73.48 & 59.05 & 33.36 & 84.00 & 87.00 & 68.71 & 23.76\% \\
 & P & 34.81 & 30.50 & 24.33 & 13.79 & 28.97 & 28.43 & 19.71 & 33.76 & 31.67 & 26.54 & 11.28 & \textbf{27.86} & 29.43 & 17.00 & 66.69\% \\
 & L & \textbf{31.73} & \textbf{29.21} & \textbf{23.17} & \textbf{13.01} & \textbf{27.21} & \textbf{27.29} & \textbf{18.00} & \textbf{32.07} & \textbf{29.60} & \textbf{25.18} & \textbf{10.35} & \textbf{27.86} & \textbf{28.86} & \textbf{16.14} & \textbf{68.4}7\%
 \\ \hline
\end{tabular}} 
\vspace{0.2em}
\caption{The attack results with different relationships:
Original Results (O), Label-wise (C), Pair-wise (P), and List-wise (L).
We evaluate the performance with six retrieval models on four evaluated datasets.
The $\mathcal{R}$Oxford5k along with $\mathcal{R}$Paris6k is annotated with three protocol setups: Easy (E), Medium (M), Hard (H).
Lower \emph{m}AP or \emph{m}P@10 and higher mDR(mean dropping rate) mean better performance in attack.
}\label{tb_total_result}
\vspace{-1em}
\end{table*}

\subsection{Random Resizing}\label{resizing}
Unlike classification models, where input images are cropped and padded to a fixed size, retrieval model can accept inputs at different scales.
Therefore, resizing is a mean for defense attack~\cite{xie2017mitigating}, which not only affects the retrieval performance, but also influences the attack quality.

To make the proposed universal perturbation be suitable for different scales, a random resizing process $R_I(\cdot)$ is employed, which resizes the original input image $x$ with size $W \times H \times 3$ to a new image $R_I(x)$ with random size $W^{'} \times H^{'} \times 3$.
Note that, $W'$ along with $H'$ is within a specific range, and $|\frac{W^{'}}{W} - \frac{H^{'}}{H}|$ should be within a  reasonably small range to prevent image distortion.
Then, the UAP $\delta$ is resized to a new perturbation $R_P(\delta, R_I(x))$ with the same size as $R_I(x)$ to be added to the input image.

\subsection{Rank Distillation}
Above methods require accessing model parameters which is not realistic in general.
To overcome it, we propose a coarse-to-fine rank distillation method to build a substitute model.
Note that the gap between different architectures exists, and distillation can be viewed as an effective defense~\cite{papernot2017extending,papernot2016distillation} as well.
Therefore, distilling with diverse architecture may not work.
Similar to~\cite{xiao2018generating}, we assume the architecture of model is known.

Since regressing large-scale ranking indices is very computational and memory intensive,
we turn to adopt a hierarchical strategy that first considers coarse-grained subset and then focuses on fine-grained top-$k$ references.

For coarse-grained part, a subset of the entire ranking list which preserves the global ranking information for distilled model to regress is considered.
Concretely, a large ranking list is divided into many bins according to the indices, and a subset is constructed by sampling one reference from each bin. 
We optimize the distillation model on the subset to fit the ordinal relation between the corresponding bins. 
Formally, the ordinal regression objective is defined as follows:
\begin{equation}\label{eq_distillation}
    \min \sum_i \sum_{m>n} \lambda_m {[d(q_i,r_{im})-d(q_i,r_{in})+\beta]}_+,
\end{equation}
where $q_i$ is the feature from the distilled model of the $i$-th query, $r_{im}$ is the feature of the $m$-th similar reference in subset for the $i$-th query, $\lambda_m$ is the discount factor ensuring top references have more importance, and $\beta$ is the margin to avoid all features falling into a single point.

Subsequently, for fine-grained part, a refined procedure focusing on the top-$k$ references are conducted. 
We adopt the similar strategy as coarse part with decreasing arguments (\emph{e.g.} learning rate and margin), while $r_{im}$ in Eq.\,(\ref{eq_distillation}) refers to the $m$-th similar feature in top-$k$ list instead.

Then, the same attack strategy as described in Sec.\,~\ref{sub_sec:white_methd} is carried out on the distilled model, and the learned perturbation is transferred to attack the true target victim.

\subsection{The Optimization}
Since the gradient of $\delta$ can be got easily, we adopt the stochastic gradient descent with momentum~\cite{dong2018boosting} to update the perturbation vector at the $i$-th iteration:
\begin{equation}\label{Eq_up}
\begin{aligned}
g_i &= \mu \cdot g_{i-1} + \frac{\nabla \delta}{\|\nabla \delta\|_1} ,
\\
\delta_i &= \delta_{i-1} + \lambda \cdot sign(g_i),
\\
\delta_i &= \min\big(\max(-\epsilon, \delta_i), \epsilon\big),
\end{aligned}
\end{equation}
where $g_i$ is the momentum of the $i$-th iteration and $\lambda$ is the learning rate.
The clipping operation that ensures constraint $\|\delta\| \leq \epsilon$ may invalidate the update after $\delta$ reaches a constraint. 
We tackle this issue by following~\cite{Mopuri2018GeneralizableDO}, which rescales $\delta$ to half when the perturbation gets saturated.
The detailed algorithm is provided in Alg.\,\ref{alg1}.

\section{Experiments}\label{sec:exp}
\begin{table}[t]
    \footnotesize
\centering
\resizebox{0.49\textwidth}{!}{
\begin{tabular}{c|c|c|c|c|c|cc}
\hline
 & A-MAC & A-GeM & V-MAC & V-GeM & R-MAC & R-GeM \\
 \hline
    A-MAC & \textbf{48.33} & 34.94 & 13.60 & 10.78 & 8.57 & 11.27 \\
    A-GeM & 38.18 & \textbf{56.88} & 14.31 & 12.00 & 7.64 & 12.22 \\
    V-MAC & 14.68 & 15.26 & \textbf{67.96} & 60.16 & 18.46 & 19.32 \\
    V-GeM & 15.66 & 16.30 & 66.16 & \textbf{63.72} & 18.24 & 19.87 \\
    R-MAC & 16.38 & 15.53 & 23.59 & 19.62 & 62.60 & 58.25 \\
    R-GeM & 14.27 & 14.29 & 23.94 & 22.35 & \textbf{67.91} & \textbf{68.47} \\

\hline
\end{tabular}
}
\caption{Results of transfer attack. The mean dropping rates are reported, where a larger number means better attack performance.}\label{tb_transfer}
\vspace{-1.5em}
\end{table}

In this section, we present quantitative results and analysis to evaluate the proposed attack schemes.
We train our universal perturbations on the $30k$ Structure-of-Motion Reconstruction dataset.
Two recent CNN-based image descriptors (\emph{i.e.}, MAC~\cite{razavian2016visual, tolias2015particular} and GeM~\cite{radenovic2018fine}) with three different CNN models (\eg{} AlexNet~\cite{krizhevsky2012imagenet}, VGGNet~\cite{simonyan2014very} and ResNet~\cite{he2016deep}) are used, forming six CNN models that are trained on the $120k$ Structure-of-Motion Reconstruction dataset.
We use \textit{Oxford5k} and \textit{Paris6k} with their revised versions~\cite{radenovic2018revisiting} to evaluate the attack performance.

\noindent\textbf{Training datasets.}
The \textit{SfM} dataset~\cite{schonberger2015single} consists of $7.4$ million images downloaded from Flickr. 
It contains two large-scale training sets named \textit{SfM-30k} and \textit{SfM-120k}, respectively.
We utilize K-Means clustering on 6,403 validation images from \textit{SfM-30k} to obtain the list-wise relationship, and use the clustering index as pseudo-label to train a classification model to obtain the  label-wise relationship. 
Our universal perturbations are trained on 1,691 query images from the \textit{SfM-30k}.

\noindent\textbf{Test Datasets.}
The \textit{Oxford5k} dataset~\cite{radenovic2018revisiting} consists of 5,062 images and the collection has been manually annotated to generate a comprehensive ground truth for 11 different landmarks, each of which is represented by 5 possible queries.
Similar to \textit{Oxford5k}, the \textit{Paris6k} dataset~\cite{radenovic2018revisiting} consists of 6,412 images with 55 queries.
Recently, Radenovi´c \emph{et al.}~\cite{radenovic2018revisiting} have revisited these two datasets to revise the annotation error, the size of the dataset, and the level of challenge.
The \textit{Revisited Oxford5k} and \textit{Revisited Paris6k} datasets are referred as $\mathcal{R}$Oxford5k and $\mathcal{R}$Paris6k, respectively.
We report our results on both the original and revisited datasets.

\noindent\textbf{Visual Features.}
For CNN-based image representation, we use AlexNet (A)~\cite{krizhevsky2012imagenet}, VGG-16 (V)~\cite{simonyan2014very} and ResNet101 (R)~\cite{he2016deep} pre-trained on ImageNet~\cite{deng2009imagenet} as our base models to fine-tune the CNN models on the \textit{SfM-120k} dataset.
For the fine-tuned features, we consider two cutting-edge features, \emph{i.e.}, the generalized mean-pooling (GeM)~\cite{radenovic2018fine} and the max-pooling (MAC)~\cite{razavian2016visual, tolias2015particular}.
As a result, we obtain a total of 6 features to evaluate the attack performance, termed as \textit{A-GeM}, \textit{V-GeM}, \textit{R-GeM}, \textit{A-MAC}, \textit{V-MAC} and \textit{R-MAC}.

\noindent\textbf{Evaluation Metrics.}
To measure the performance of universal perturbation for retrieval, we mainly consider three evaluation metrics, \emph{i.e.} ,\emph{m}AP, \emph{m}P@10, and the fooling rate.
Unlike classification, the fooling rate of top-1 label prediction can not be computed directly for image retrieval.
Therefore, we define a new metric to evaluate the fooling rate for retrieval, termed dropping rate (DR) as follows:
\begin{equation}\label{fooling_rate}
DR(M, x, \hat{x}) = \frac{M(x)-M(\hat{x})}{M(x)} \times 100\%,
\end{equation}
where $\hat{x}$ is an adversarial example of the original feature $x$, and $M$ is the metric used in retrieval such as \emph{m}AP.
Dropping rate characterizes the attack performance by measuring the performance degeneration of retrieval systems.
The higher the dropping rate is, the more successful the attack is.

\subsection{Results of UAP Attack}
We evaluate the performance of six state-of-the-art deep visual representations against universal adversarial perturbation, the quantitative results of mean DR, \emph{m}AP and \emph{m}P@10 are shown in Tab.\,\ref{tb_total_result}.
Poor dropping rates (except ones for VGG16) prove limited ability of UAPs against classification on retrieval.
Although they achieve considerable results for VGG16, they are still worse than our proposed methods.
Clearly, for all deep visual features, all kinds of our universal perturbations achieve very high dropping rates on the validation set.
Most of them achieve a dropping rate of more than $50\%$, which means that most relevant images will not be returned on top of the ranking list.
Specifically, the universal perturbations computed for V-MAC and R-GeM achieve nearly $68\%$ dropping rate.
Notably, list-wise relationship plays an important role in generating universal perturbations. 
We owe it to more ranking information employed during optimization.
We conclude that both the pair-wise and list-wise relationships are suitable for universal perturbation generation, and list-wise relationship achieves better performance.

\begin{figure*}
    \begin{center}
    \begin{minipage}[t]{0.5\linewidth}
    \centerline{
    \subfigure[Oxford5k]{
    \includegraphics[width=\linewidth]{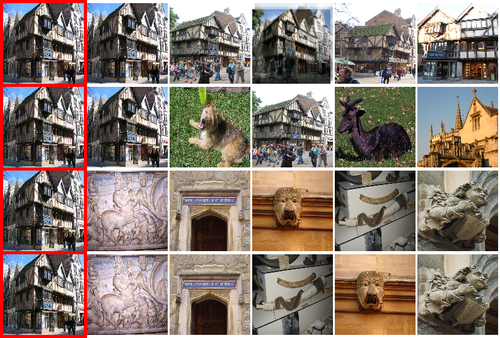}}
    \subfigure[Paris6k]{
    \includegraphics[width=\linewidth]{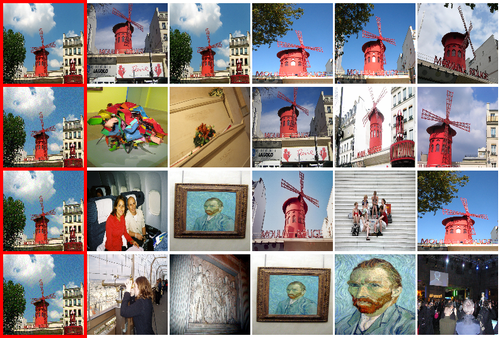}}
    }
    \end{minipage}
    \end{center}
    \vspace{-1em}
    \caption{The visualization results on \textit{Oxford5k} and \textit{Paris6K} for ResNet101-GeM. All the images in red box are the queries, and the retrieved pictures are sorted from left to right. The 4 rows show the retrieval results by using the original images and perturbed images via the label-wise relationship, pair-wise relationship and list-wise relationship, respectively. (Best viewed in color.) }
~\label{fg_attack_result}
\vspace{-2em}
\end{figure*}

\subsection{Results of Transfer Attack}
\begin{table}[t]
\centering
\begin{tabular}{c|c|c|cc}
\hline
 & Random & Pre-trained & Distillation  \\ \hline
    A-GeM & 5.53\% & 32.98\% & 39.72\% \\
    V-GeM & 1.66\% & 28.85\% & 44.68\% \\ \hline
\end{tabular}
\caption{Results about distillation attack. Random refers to Perturbations on randomly initialized model, pre-trained means the one from model trained on the ImageNet dataset, and distillation is obtained via distilled model.}\label{tb_distillation}
\vspace{-1em}
\end{table}

As mentioned in Sec.\,\ref{sec:R}, transfer attack is to fool models or dataset with a perturbation generated on another model or dataset.
Tab.\,\ref{tb_transfer} shows the results about the transfer attack across different visual features, in which we report the \emph{m}DR calculated on all four evaluation datasets.
Each row in Tab.\,\ref{tb_transfer} shows the \emph{m}DRs for perturbation crafted by a given model, and each column shows the transfer dropping rates on the target model.
The universal perturbation is trained on one architecture (\emph{e.g.,} V-GeM), whose attack ability is evaluated to fool the retrieval system based on the other deep features (\emph{e.g.,} R-MAC or V-MAC\footnote{We consider that different CNN architecture with the same pooling method as different features.}).
It is interesting to find that universal perturbations generated from the same network architecture can be transferred well to related models with different pooling methods.

We also measure the power of distillation in Tab.\,\ref{tb_distillation} for the case that the architecture is known beforehand.
It's clear that perturbations from randomly initialized models make no sense in spite of using the same architecture.
As all the retrieval models are fine-tuned from the ImageNet pre-trained models, perturbations generated from pre-trained models achieve considerable results compared with transfer attack from other architectures in Tab.\,\ref{tb_transfer}.
However, perturbations from distilled model overmatch ones from pre-trained models by at least $6\%$, showing the power of ranking distillation.
We conclude that our proposed ranking distillation attack is practical, when the model parameters can not be touched.

\begin{figure}[!t]
    \begin{center}
 \includegraphics[width=\linewidth]{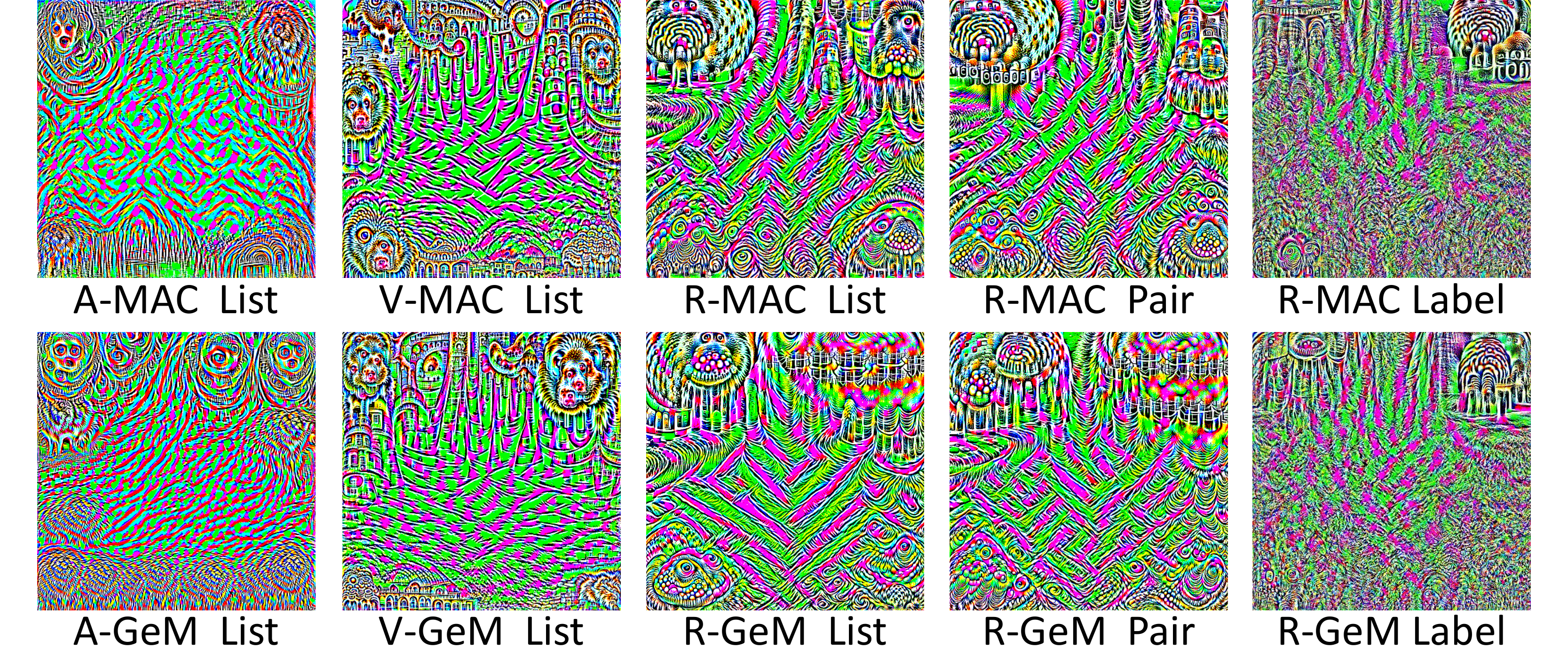}
    \end{center}
\vspace{-1em}
\caption{Universal adversarial perturbations crafted by the proposed method for multiple architectures trained on SfM. Corresponding features and deep architectures are mentioned below each image. (Best viewed in color and zoom in.) }
~\label{fg_noise}
\vspace{-2em}
\end{figure}

\begin{table}[]
  \footnotesize
\centering

\resizebox{0.49\textwidth}{!}{
\begin{tabular}{c|cccccc}
\hline
  Range & \begin{tabular}[c]{@{}c@{}}[362,\\ \ \ 362]\end{tabular}
          & \begin{tabular}[c]{@{}c@{}}[1024,\\ \ \ 1024]\end{tabular}
          & \begin{tabular}[c]{@{}c@{}}[128,\\ \ \ \ 1024]\end{tabular}
          & \begin{tabular}[c]{@{}c@{}}[256,\\ \ \ \ 1024]\end{tabular}
          & \begin{tabular}[c]{@{}c@{}}[512, \\ \ \ \ 1024]\end{tabular}
          & \begin{tabular}[c]{@{}c@{}}[768,\\ \ \ \ 1024]\end{tabular} \\ \hline

\hline
  A-GeM & 16.89\% & 24.69\% & 53.21\% & 56.88\% & 51.41\% & 39.21\% \\
  V-GeM & 25.87\% & 30.42\% & 61.93\% & 63.72\% & 55.02\% & 42.08\% \\
\hline
\end{tabular}
}
\caption{The effect of resizing in attack.}\label{tb_resizing}
\vspace{-1.5em}
\end{table}

\subsection{On the Effect of Resizing}
As mentioned before, the retrieval system can accept various size of input image, which inspire us to investigate the effect of resizing when attacking the systems.
Quantitative results are shown in Tab.\,\ref{tb_resizing}.
We first set the resizing scale to a fixed $362\times 362$ and $1024 \times 1024$,
considering that $362 \times 362$ is the scale used to training the retrieval model.
The dropping rates for A-GeM and V-GeM are lower than half of our multi-scale random resizing method.
Finally, we evaluate the influence of the range for our multi-scale random resizing and observe that too broad or narrow range damages attack performance.

\subsection{Visualization}\label{vis}
Fig.\,\ref{fg_attack_result} shows the retrieval results for R-GeM features from the \textit{Oxbuild5k} and \textit{Paris6K} evaluation set.
In details, to attack the label-wise relationship, the model aims to learn the perturbation to push the original image to other categories.
In the second row, we observe that the top 5 retrieved images are relevant to the category of dogs, instead of the true category of building.
This phenomenon exists for pair-wise relationship and list-wise relationship that both pursues the farthest landmark to some degree, \emph{e.g.,} most retrieved images are relevant about sculptures or oil paintings.
Note that, retrieved images for pair-wise relationship and list-wise are similar since list-wise relationship includes pair-wise information.

We then visualize the perturbations that are trained from different models in Fig.\,\ref{fg_noise}.
Perturbations in the first row are generated from MAC pooling and the ones in second row are from GeM pooling.
The first three perturbation each row generated from different networks show large difference, while perturbations from same column share similar appearances.
This is consistent with the result of transfer attack.
Besides, perturbations crafted from pair-wise relationship and list-wise relationship are more similar than the one from label-wise, which may also indicate the gap between attack of classification and retrieval.

\subsection{The Real-world System Attack}
\begin{figure}[!t]
 \centering
 \includegraphics[width=3.3in]{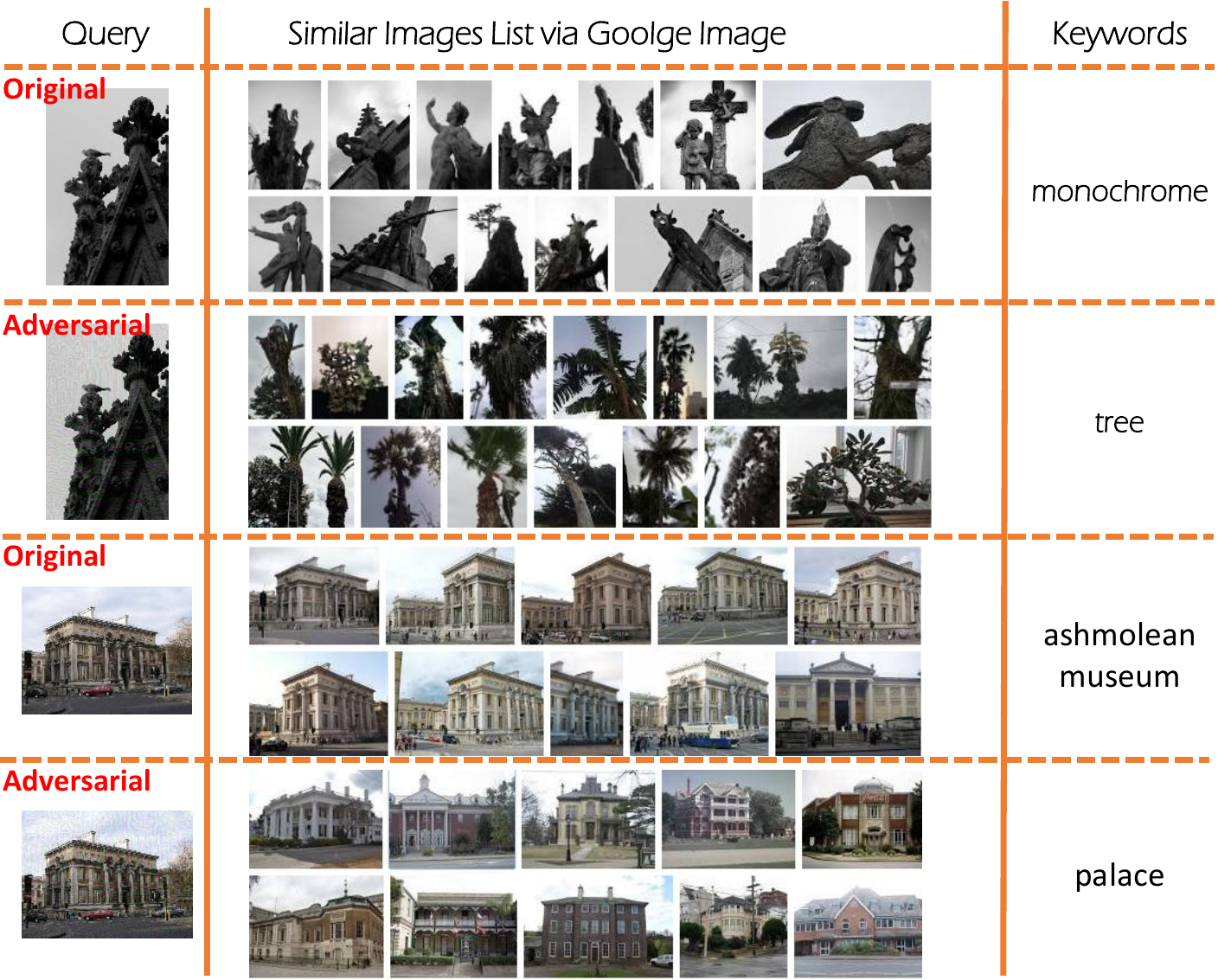}
 \caption{Example retrieval results on Google Images. 
     The odd rows and even rows show the images retrieved by original query and the corrupted ones by our universal perturbation, respectively.
 The predicted keywords via Google Image are also given.}~\label{fg_google}
\vspace{-2em}
 \end{figure}

Fig.\,\ref{fg_google} shows the attack results on a real-world image retrieval system, \emph{i.e.}, Google Image.
The even rows show the perturbed images along with the retrieved images and the predicted keywords provided by Google Image, which are completely different from the original ones at the odd rows.
For example, the original input is categorized to monochrome, while the adversarial example changes to be tree.
Note that it is unable to quantize the \emph{m}AP drop due to the lack of ground truth ranking list.
Therefore, we quantize how often the retrieved images from clean query are absent in the retrieved list of the corrupted one for 100 images randomly sampling from \textit{Oxbuild5k} and \textit{Paris6K} datasets.
For this metric, our model has a 62.85\% absent rate achieved.
The attack results have demonstrated that the proposed method can generate universal perturbations to fool the real-world search engine.

\section{Conclusion}\label{sec:cc}

In this paper, we are the first to propose a set of universal attack methods against image retrieval.
We mainly focus on attacking the point-wise, pair-wise, and list-wise neighborhood relationships.
We further analyze the impact of resizing operations in generating universal perturbation in details,
and employ a multi-scale random resizing method to improve the success rate of the above attack schemes.
A coarse-to-fine distillation strategy is also been proposed for black-box attack.
We evaluate our proposed method on widely-used image retrieval datasets, \emph{i.e.}, \textit{Oxford5k}, and \textit{Paris6K}, in which our method shows high attack performance that leads to a large retrieval metrics drop in a serial of models.
Finally, we also attack the real-world system, \emph{i.e.}, Google Images, which further demonstrates the efficacy of our methods.
Last but not least, our work can therefore serve as an inspiration in designing more robust and secure retrieval models against the proposed attack schemes.

\paragraph{Acknowledgements.}
\footnotesize{
  This work is supported by the National Key R\&D Program (No.2017YFC0113000 and No.2016YFB1001503), Nature Science Foundation of China (No.U1705262, No.61772443, and No.61572410), Scientific Research Project of National Language Committee of China (No.YB135-49), and Nature Science Foundation of Fujian Province, China (No.2017J01125 and No.2018J01106).
}

{\small
\bibliographystyle{ieee_fullname}
\bibliography{egbib}
}

\end{document}